\def\year{2020}\relax
\documentclass[letterpaper]{article} % DO NOT CHANGE THIS
\usepackage{aaai20}  % DO NOT CHANGE THIS
\usepackage{times}  % DO NOT CHANGE THIS
\usepackage{helvet} % DO NOT CHANGE THIS
\usepackage{courier}  % DO NOT CHANGE THIS
\usepackage[hyphens]{url}  % DO NOT CHANGE THIS
\usepackage{graphicx} % DO NOT CHANGE THIS
\usepackage{graphicx}  % DO NOT CHANGE THIS
\usepackage{latexsym}
\usepackage[utf8]{inputenc}

\usepackage{amsmath}
\usepackage{booktabs}
\usepackage{algorithm}
\usepackage{algorithmic}
\usepackage{multirow}
\usepackage{pifont}
\usepackage{bm}

\title{Rule-Guided Compositional Representation Learning on Knowledge Graphs}
\author{
Guanglin Niu,\textsuperscript{\rm 1}
Yongfei Zhang,\textsuperscript{\rm 1,2}
Bo Li\thanks{Corresponding author.},\textsuperscript{\rm 1,2}
Peng Cui,\textsuperscript{\rm 3}
Si Liu,\textsuperscript{\rm 1}
Jingyang Li,\textsuperscript{\rm 1}
Xiaowei Zhang\textsuperscript{\rm 4}
\\ % All authors must be in the same font size and format. Use \Large and \textbf to achieve this result when breaking a line
\textsuperscript{\rm 1}Beijing Key Laboratory of Digital Media, School of Computer Science and Engineering, Beihang University,\\
\textsuperscript{\rm 2}State Key Laboratory of Virtual Reality Technology and Systems, Beihang University,\\
\textsuperscript{\rm 3}Department of Computer Science and Technology, Tsinghua University,\\
\textsuperscript{\rm 4}College of Computer Science and Technology, Qingdao University
\\ %If you have multiple authors and multiple affiliations
beihangngl@buaa.edu.cn, yfzhang@buaa.edu.cn, boli@buaa.edu.cn, cuip@tsinghua.edu.cn\\
liusi@buaa.edu.cn, lijingyang@buaa.edu.cn, xiaowei19870119@sina.com% email address must be in roman text type, not monospace or sans serif
}

\begin{document}

\maketitle

\begin{abstract}
Representation learning on a knowledge graph (KG) is to embed entities and relations of a KG into low-dimensional continuous vector spaces. Early KG embedding methods only pay attention to structured information encoded in triples, which would cause limited performance due to the structure sparseness of KGs. Some recent attempts consider paths information to expand the structure of KGs but lack explainability in the process of obtaining the path representations. In this paper, we propose a novel Rule and Path-based Joint Embedding (RPJE) scheme, which takes full advantage of the explainability and accuracy of logic rules, the generalization of KG embedding as well as the supplementary semantic structure of paths. Specifically, logic rules of different lengths (the number of relations in rule body) in the form of Horn clauses are first mined from the KG and elaborately encoded for representation learning. Then, the rules of length 2 are applied to compose paths accurately while the rules of length 1 are explicitly employed to create semantic associations among relations and constrain relation embeddings. Moreover, the confidence level of each rule is also considered in optimization to guarantee the availability of applying the rule to representation learning. Extensive experimental results illustrate that RPJE outperforms other state-of-the-art baselines on KG completion task, which also demonstrate the superiority of utilizing logic rules as well as paths for improving the accuracy and explainability of representation learning.
\end{abstract}

\section{Introduction}
Knowledge graphs (KGs) such as Freebase \cite{BGF:Freebase}, DBpedia \cite{Lehmann:dbpedia} and NELL \cite{Mitchell:nell} are knowledge bases which store factual triples consisting of entities with their relations. They have achieved rapid development and extensive applications for various research fields, such as zero-shot recognition \cite{Wang:CVPR}, question answering \cite{Hao:KBQA} and recommender systems \cite{Zhang:recommender}.

\begin{figure}
	\includegraphics[width=1\columnwidth]{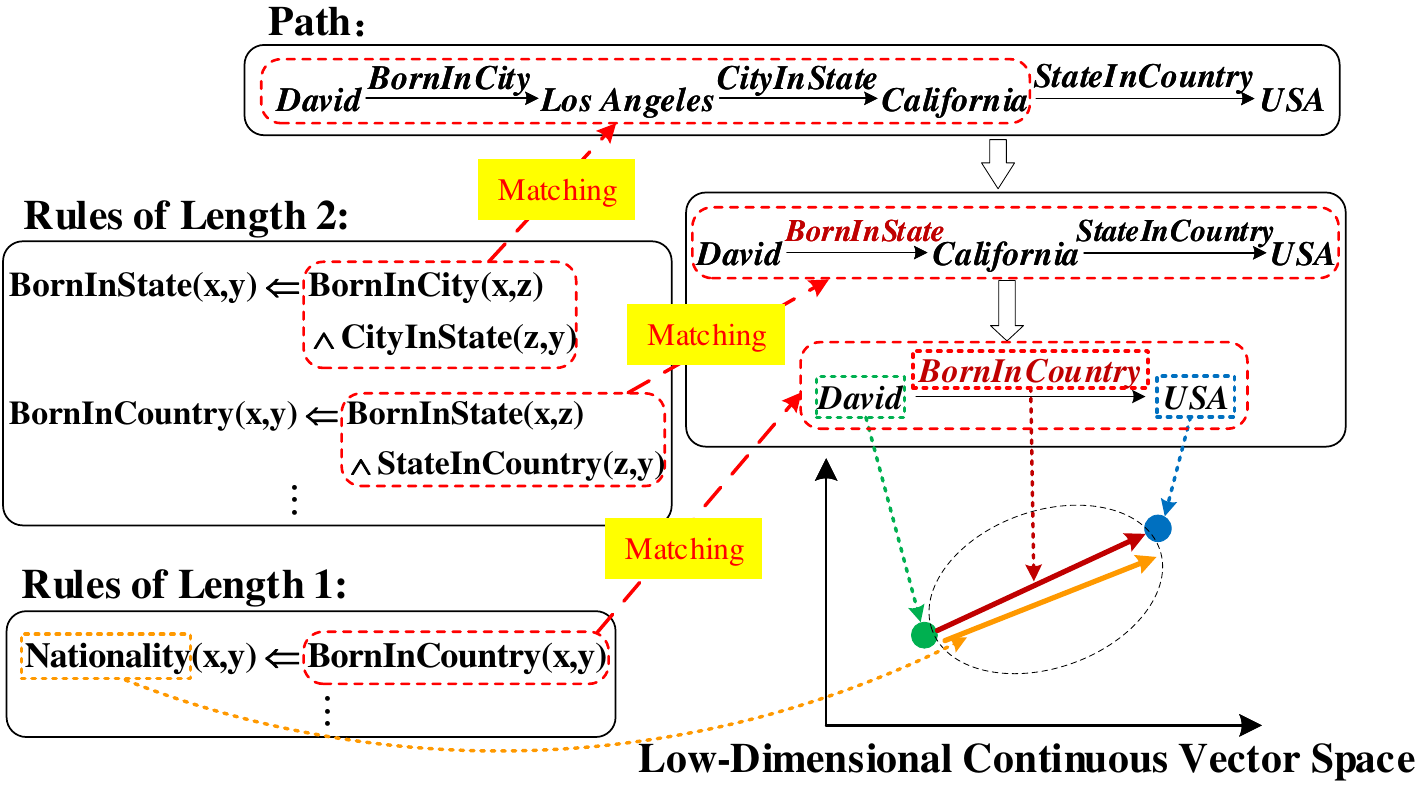}
	\caption{A motivating example of our proposed approach based on Horn rules and a path.}
	\label{figure1}
%	\vspace{-0.3cm}
\end{figure}

Typical KGs contain billions of triples but remain to be incomplete. Specifically, 75\% of 3 million person entities miss a nationality in Freebase \cite{West:WWW} and 60\% of person entities do not have a place of birth in DBpedia \cite{Krompab:ISWC}. Thus, it is hard to further introduce KGs into some applications, such as no correct answers for question answering systems based on incomplete KGs. And the symbolic nature of triple facts in the form of (head entity, relation, tail entity) makes it challenging to expand large scale KGs.

In recent years, representation learning on KGs (or known as KG embedding) such as TransE \cite{Bordes:TransE}, TransH \cite{Wang:TransH} and TransR \cite{Lin-a:TransR} has become popular which intends to embed entities and relations of KGs into a continuous vector space while retaining the inherent structure and latent semantic information of KGs \cite{Qang:Survey}. This could benefit a lot for large scale KG completion.

The above embedding models purely consider the single triples. In fact, multi-step paths in KGs always play a pivotal role in providing extra relationships between entity pairs. For instance, we can utilize a path in the KG \small${David \xrightarrow{BornInState} California \xrightarrow{StateInCountry} USA}$ \normalsize to expand the knowledge for its corresponding triple fact \small$(David, Nationality, USA)$. \normalsize Both Lin et al. \shortcite{Lin-b:PTransE} and Guu et al. \shortcite{Kelvin:Traversing} succeeded to learn the entity and relation embeddings on paths. In their work, the relation embeddings are initialized randomly and the path representations are composed via addition, multiplication or Recurrent Neural Networks (RNN) of the relations along the paths. Since the path representations are achieved purely based on numerical calculations in a latent space, such approaches will cause error propagation and limited accuracy of the paths embeddings and further affect the whole representation learning.

An effective way to apply the extra semantic information in KG embedding is to employ the logic rules in view of their accuracy and explainability. KALE \cite{Shu-Guo:Jointly} and RUGE \cite{Shu-Guo-2:RUGE} both convert the rules into complex formulae modeled by t-norm fuzzy logics to transfer the knowledge in rules into the learned embeddings. Nevertheless, logic rules could maintain their original explainability only in the symbolic form and such rules actually focus on the semantic associations as well as the constraints of various relations, which have not been well exploited for the triples and paths in KG embedding.

This paper proposes a novel rule-guided compositional representation learning approach named Rule and Path-based Joint Embedding (RPJE), using Horn rules to compose paths and associate relations in the semantic level to improve the precision of learning KG embeddings on paths and enhance the explainability of our representation learning. In allusion to the path as shown in Figure \ref{figure1}, the rule bodies in two Horn rules of length 2
\small${borninstate(x,y)\Leftarrow bornincity(x,z)\wedge cityinstate(z,y)}$ and
\small${bornincounrty(x,y)\Leftarrow borninstate(x,z)\wedge stateincountry}$
\small${(z,y)}$ \normalsize
are respectively matched with two segments of the path, which could be applied to iterately compose the entire path as a straightforward triple \small${(David}$${, BornInCountry, USA)}$. \normalsize Then, the rule body of length 1 \small${Nationality(x,y) \Leftarrow BornInCountry(x,y) (0.9)}$ \normalsize is matched with the relation \small$BornInCountry$. \normalsize Therefore, two relation embeddings denoting \small$Nationality$ \normalsize and \small$BornInCountry$ \normalsize are further constrained to be closer in the latent space with confidence level 0.9.

In experiments, we evaluate the proposed model RPJE on four benchmark datasets of FB15K, FB15K-237, WN18, and NELL-995. Experimental results illustrate that our approach achieves superior performances on KG completion task and significantly outperforms state-of-the-art baselines, which verifies the capability of combining rules with paths in KG embedding. Our main contributions of this work are:

\begin{itemize}

\item	To the best of our knowledge, this is the first attempt to integrate logic rules with paths for KG embedding, endowing our model with both the explainability from semantic level and the generalization from data level.

\item	Our proposed model RPJE considers the various types of rules to inject prior knowledge into KG embedding. It can use the encoded length-2 rules to start paths composition rather than with randomly initialized vectors for obtaining accurate path representations. And the semantic associations among relations could be created by length-1 rules.
%which inject the prior knowledge into compositional representation learning.

\item	We conduct extensive experiments on KG completion and our model achieves promising performances. The influence of different confidence thresholds of rules demonstrates that considering confidence of rules in our model guarantees the effectiveness of using rules and could achieve good robustness to various confidence thresholds.
%And the performance by the different confidence thresholds of rules demonstrates that considering confidence of rules in our model could guarantee the effectiveness of using rules with various confidence levels. and have good robustness to confidence thresholds.

\end{itemize}

\section{Related Work}

\textbf{KG embedding models:}
In recent years, many works have been done to learn distributed representations for entities and relations in KGs, which fall into three major categories: (1) Translational distance model. Inspired by the translation invariant principle from word embedding \cite{Mikolov:WordEmbedding}, TransE \cite{Bordes:TransE} regards relations as translating operations between head and tail entities, i.e., the formula $h+r \approx t$ should be satisfied when triple $(h,r,t)$ holds. (2) Tensor decomposition model. DistMult \cite{Yang:ICLR} and ComplEx \cite{Trouillon:ComplEx} both utilize tensor decomposition to represent each relation as a matrix and each entity as a vector. 3) Neural networks model. In NTN \cite{Socher:NTN}, a 3-way tensor and two transfer matrices are encoded into multilayer neural network. Among these methods, TransE and plenty of its variants TransH \cite{Wang:TransH}, TransR \cite{Lin-a:TransR} and TransG \cite{Xiao:TransG} have become promising approaches for successfully capturing the semantics of KG symbols. However, the above methods merely consider the facts immediately observed in KGs and ignore extra prior knowledge to enhance KG embedding.

\textbf{Path enhanced models:}
Paths existing in KGs have gained more attentions to be combined with KG embedding because multi-hop paths could provide relationships between seemingly unconnected entities in KGs. Path Ranking Algorithm (PRA) \cite{Lao:PRA} is one of the early studies which searches paths by random walk in KGs and regards the paths as features for a per-target relation binary classifier. Neelakantan et al. \shortcite{Neelakantan:Compositional-vector-space} proposed a compositional vector space model with a recurrent neural network to model relational paths on knowledge graph completion. Guu et al. \shortcite{Kelvin:Traversing} introduced additive and multiplicative interactions between relation matrices in the path. Lin et al. \shortcite{Lin-b:PTransE} proposed PTransE to obtain the path embeddings by composing all the relations in each path. DPTransE \cite{DPTransE} jointly builds interactions between the latent features and graph features of KGs to offer precise and discriminative embedding. However, all these techniques obtain the path representations via calculating relation embeddings along the paths, which would cause limited accuracy and lack explainability.

\textbf{Rule extraction and rule enhanced models:}
Logic rules are explainable and contain rich semantic information, which have shown the power in knowledge inference. The Inductive Logic Programming algorithms such as XAIL are available to learn FOL or even ASP-style rules. However, it is difficult to mine rules from KGs with ILP algorithms due to open world assumption of KGs (absent information cannot be taken as counterexamples). Therefore, several rule mining methods have been developed to extract rules efficiently from large scale KGs, including AMIE \cite{AMIE:WWW}, AMIE+ \cite{Galarrage:AMIE}, RLvLR \cite{Omran:IJCAI} and CARL \cite{CARL}.

Based on the valid rules, some studies integrate logical rules in deep neural networks (DNN) and show impressive results in sentence sentiment analysis and named entity recognition \cite{HarnessingDNN,DNNwithknowledge}. Markov logic network (MLN) \cite{MLN} combines first-order logic with probabilistic graphical models for reasoning but it is inefficient to infer via MLN approach and the performance is limited due to some triples cannot be discovered by any rules. Besides, some works \cite{Daria:1,Daria:2} obtain the answer set semantics by directly using horn rules on KGs for KG completion but lack high generalization and the main usage of these models is learning rules.

To improve both the precision and generalization of KG completion, some of the recent researches attempt to incorporate the rules into KG embedding models \cite{Wang:KBCR}. Minervini et al. \shortcite{Complex-R} imposed the equivalence and inversion constraints on the relation embeddings. But this approach considers only two types of constraints between relations rather than general rules, which might not always be available for any KG. In both KALE \cite{Shu-Guo:Jointly} and RUGE \cite{Shu-Guo-2:RUGE}, the triples are represented as atoms and the rules are modeled by t-norm fuzzy logic for being converted into complex formulae formed by atoms with logical connectives. However, the above two methods quantify the rules in embedding which would decrease the explainability and accuracy of rules. By eliminating the complex process of modeling rules to be formulae, we explicitly and immediately employ the Horn rules to deduce the path embeddings and create the semantic associations between relations.

\begin{figure*}
	\centering
	\includegraphics[scale=0.86]{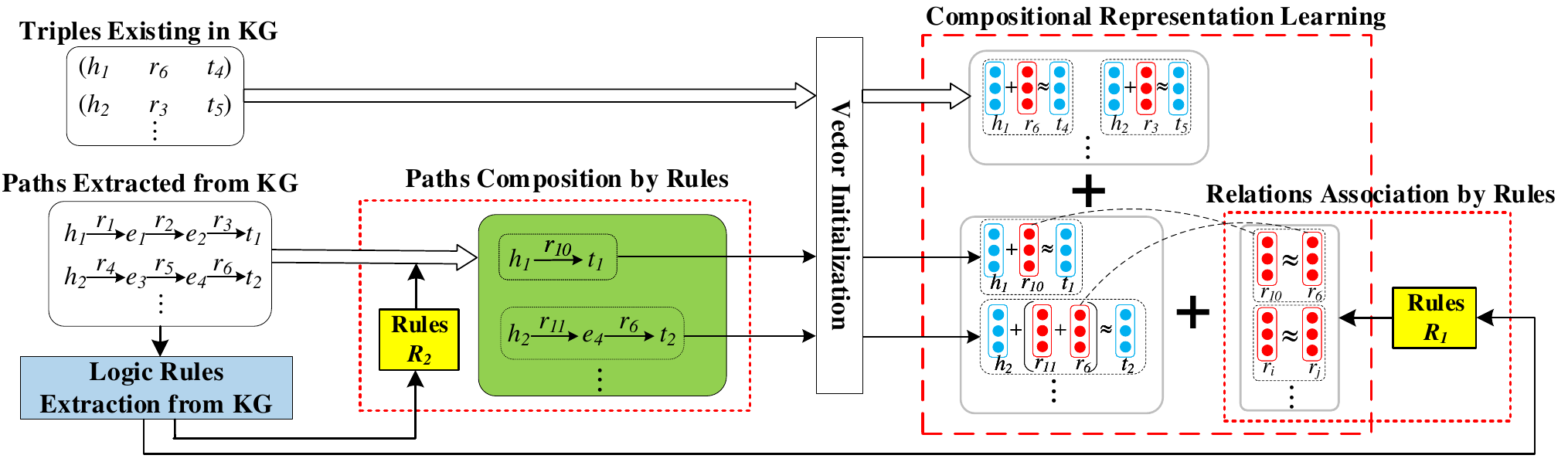}
	\caption{Overall architecture of our model.}
	\label{figure2}
\end{figure*}

\section{Methodology}

We attempt to integrate paths with logic rules to provide more semantic information in our model. The overall framework of the proposed scheme is shown in Figure \ref{figure2}. Firstly, we extract the paths and mine the Horn rules from KG, where the rules of length 1 and 2 are denoted as Rules $R_1$ and Rules $R_2$, respectively (§\ref{section3.1}). Then, we apply Rules $R_2$ to iteratively compose paths and Rules $R_1$ to create the semantic associations of some relation pairs (§\ref{section3.2}). Furthermore, vector initialization is used to transform the entities and relations in symbolic space into the vector space for training the KG embeddings. Finally, compositional representation learning is implemented for optimizing objective specific to triples, paths and associated relations pairs (§\ref{section3.3}, §\ref{section3.4}).

\subsection{Logic Rules Extraction from KG}
\label{section3.1}

Horn rules could be mined automatically by any KG rule exaction algorithm or tool. In this paper, we first mine rules together with their confidence levels denoted as $\mu \in [0,1]$ from KGs. And a rule with higher confidence level has higher possibility to hold. We limit the maximum length of rules to 2 for the efficiency of mining valid rules. Thus, rules are classified into two types according to their length: (1)\textbf{ Rules \bm{$R_1$}}. The set of length-1 rules is denoted as Rules $R_1$, which associating two relations in rule body and rule head. (2)\textbf{ Rules \bm{$R_2$}}. The set of rules with length 2 is denoted as $R_2$, which could be utilized to compose paths. Some examples of Rules $R_1$ and Rules $R_2$ are provided in Table \ref{table1}.

\begin{table*}
\centering
\resizebox{\textwidth}{!}{
\begin{tabular}{l}
\toprule
Rules $R_1$ with confidence levels\\
\midrule
$concept:citylocatedincountry(a,b) \Leftarrow concept:citycapitalofcountry(a,b)$ 0.86 (extracted from NELL-995)\\
$/location/location/people\_born\_here(a,b) \Leftarrow /people/person/place\_of\_birth^{-1}(a,b)$ 1 (extracted from FB15K)\\
\toprule
Rules $R_2$ with confidence levels\\
\midrule
$concept:parentofperson(a,b) \Leftarrow concept:hasspouse(a,e) \wedge concept:fatherofperson(e,b)$ 1 (extracted from NELL-995)\\
$/film/language(a,b) \Leftarrow /film/directed\_by(a,e) \wedge /person/language(e,b)$ 0.81 (extracted from FB15K)\\
\bottomrule
\end{tabular}}
\caption{A few examples of rules mined from FB15K and NELL-995. The superscript "-1" means inverse relation.}
\label{table1}
\end{table*}

\textbf{Remark}: The inverse version of each relation is always added in path-based approaches to constrain each path along one direction and improve the graph connectivity \cite{DPTransE}. Therefore, given a triple $(h,r,t)$, a reconstructed triple $(t,r^{-1},h)$ is defined to express the inverse relationship $r^{-1}$ between entity $t$ and entity $h$.

\subsection{Rules Employment for Compositional Representation Learning}
\label{section3.2}
The Horn rules extracted from KGs could be utilized in two modules for compositional representation learning, including paths composition by Rules $R_2$ and relation pairs association by Rules $R_1$.
\subsubsection{Paths Composition by Rules \bm{$R_2$}.}
We first implement paths extraction procedure by PTransE \cite{Lin-b:PTransE} on KGs, where each path $p$ is extracted together with its reliability which is achieved by the path-constraint resource allocation mechanism and denoted as $R(p\vert h,t)$ between an entity pair $(h,t)$. We generate each path set $P(h,t)$ by selecting the paths between the entity pair $(h,t)$ with their reliability over 0.01. Specifically, it is essential to form a sequential path by atoms of each rule body in Rules $R_2$ for composing paths. A chain rule is further defined as the rule which the entity pair linked by the chain in the rule body is also connected by the relation in the rule head. However, the rules mined by most of the existing open-source rule mining systems could not be directly utilized because these rules are not chain rules. Therefore, we should encode each rule to form a directed path of its rule body (removing some of the rules could not be converted as this formalization in any case). In total, there are totally 8 different types of rules conversion modes, as provided in Table \ref{table2}. Take the original rule $r_3(a,b) \Leftarrow r_1(\textbf{e},b) \wedge r_2(\textbf{e},a)$ for instance, we first convert the atom $r_2(\textbf{e},a)$ into $r_2^{-1}(a,\textbf{e})$, and then exchange two atoms in the rule body to obtain a chain rule $r_3(a,b) \Leftarrow r_2^{-1}(a,\textbf{e}) \wedge r_1(\textbf{e},b)$, which could be further abbreviated to $r_3 \Leftarrow (r_2^{-1},r_1)$. Then, a path containing a sequence of relations $r_2^{-1} \rightarrow r_1$ could be composed as $r_3$. %Take the original rule $r_3(a,b) \Leftarrow r_1(e,b) \wedge r_2(e,a)$ for example, we can replace the relation $r_2$ with its inverse relation $r_2^{-1}$ and then swap the positions of $r_1,r_2^{-1}$, so that a satisfied rule $(a,r_3,b) \Leftarrow r_2^{-1}(a,e) \wedge r_1(e,b)$ can be obtained. Next, we encode this rule as $r_3 \Leftarrow (r_2^{-1},r_1)$. In particular, we employ a hash table to store these encoded rules. The rule body is considered as key and the rule head represents the value.

\begin{table}\small
\centering
\begin{tabular}{cl}
\toprule
The original rules  & Encoded rules \\
\midrule
$r_3(a,b) \Leftarrow r_1(a,\textbf{e}) \wedge r_2(\textbf{e},b)$       & $r_3 \Leftarrow (r_1,r_2)$     \\
$r_3(a,b) \Leftarrow r_1(\textbf{e},b) \wedge r_2(a,\textbf{e})$       & $r_3 \Leftarrow (r_2,r_1)$     \\
$r_3(a,b) \Leftarrow r_1(\textbf{e},b) \wedge r_2(\textbf{e},a)$       & $r_3 \Leftarrow (r_2^{-1},r_1)$     \\
$r_3(a,b) \Leftarrow r_1(\textbf{e},a) \wedge r_2(\textbf{e},b)$       & $r_3 \Leftarrow (r_1^{-1},r_2)$     \\
$r_3(a,b) \Leftarrow r_1(a,\textbf{e}) \wedge r_2(b,\textbf{e})$       & $r_3 \Leftarrow (r_1,r_2^{-1})$     \\
$r_3(a,b) \Leftarrow r_1(b,\textbf{e}) \wedge r_2(a,\textbf{e})$       & $r_3 \Leftarrow (r_2,r_1^{-1})$     \\
$r_3(a,b) \Leftarrow r_1(\textbf{e},a) \wedge r_2(b,\textbf{e})$       & $r_3 \Leftarrow (r_1^{-1},r_2^{-1})$     \\
$r_3(a,b) \Leftarrow r_1(b,\textbf{e}) \wedge r_2(\textbf{e},a)$       & $r_3 \Leftarrow (r_2^{-1},r_1^{-1})$     \\
\bottomrule
\end{tabular}
\caption{The list of conversion mode for Rules $R_2$. The left half are the original rules directly extracted from KG, and the right half are the encoded rules. In Table \ref{table2}, variables $a,b,e$ can be substituted by entities, where $r_1,r_2$ are denoted as relations in rule body and $r_3$ is the relation in rule head.}
\label{table2}
\end{table}

To make the best of encoded rules, we should traverse the paths and conduct the composition operation iteratively in the semantic level until no relation could be composed by rules since two relations are composed every time and the composition result might be composed in the next step. Considering two types of scenarios in practical paths composition procedure: (1) The optimal scenario that all of the relations in a path could be iteratively composed by Rules $R_2$ and finally joined together as a single relation between entity pair. (2) The general scenario that some relations are unable to be composed based on Rules $R_2$, we will adopt the numerical operation such as addition for the embeddings of these relations. Besides, in allusion to the situation that more than single rule could be matched in the path simultaneously, such as two rules $U(a,b) \Leftarrow R(a,c) \wedge T(c,b)$ as well as $V(a,b) \Leftarrow R(a,c) \wedge T(c,b)$ are both activated, the rule with the highest confidence should be selected to compose the path. Specifically, we define the path composition result via the above procedure as $C(p)$ which is also denoted as the path embedding of the path {p}.

\subsubsection{Relations Association by Rules \bm{$R_1$}.}

On account of the Rules $R_1$, where a relation $r_1$ may have more semantic similarity with its directly implicating relation $r_2$ when the rule $\forall x,y:r_2(x,y) \Leftarrow r_1(x,y)$ holds. The rules in the form of $(a,r_2,b) \Leftarrow (b,r_1,a)$ need to be encoded as $(a,r_2,b) \Leftarrow (a,r_1^{-1},b)$ for representation learning. Hence, in the training process, the embeddings denoting a pair of relations which appear simultaneously in Rules $R_1$ should be constrained to be closer than the embeddings of two relations that mismatch any rule.
%\vspace{-0.15cm}

\subsection{Compositional Representation Modeling}
\label{section3.3}

Along with the strategy of translation-based algorithms, for each triple $(h,r,t)$, we define three energy functions to respectively model correlations with the direct triple along with the typical translation-based methods, the path pair using Rules $R_2$ and the relation pair employing Rules $R_1$:
\begin{align}
 E_1(h,r,t) &= \Vert \textbf{h}+\textbf{r}-\textbf{t} \Vert \label{eq2}\\
 E_2(p,r) &= R(p\vert h,t)(\prod_{\mu_i \in B(p)} \mu_i) \Vert C(p)-\textbf{r} \Vert \label{eq3}\\
 E_3(r,r_e) &= \Vert \textbf{r}-\textbf{r}_e \Vert \label{eq4}
\end{align}
where $E_1(h,r,t)$ is defined with the less score if triple $(h,r,t)$ holds. $E_2(p,r)$ denotes the energy function evaluating the similarity between path $p$ and relation $r$, and $R(p\vert h,t)$ represents the reliability of the path $p$ from the entity pair $(h,t)$ and is calculated the same as in PTransE \cite{Lin-b:PTransE}. $\textbf{h}$, $\textbf{r}$ and $\textbf{t}$ are the embeddings of head entity, relation and tail entity, respectively. $C(p)$ denotes the composition result of the path $p$, which is obtained according to the paths composition procedure explained in §\ref{section3.2}. And $B(p)=\{\mu_1,\dots,\mu_n \}$ represents the set of confidence levels corresponding to all the rules in Rules $R_2$ employed in the process of composing the path $p$. $E_3(r,r_e)$ is the energy function indicating the similarity of relation $r$ and another relation $r_e$ and should be assigned with less score if $r_e$ is the relation implicated by the relation $r$ with Rules $R_1$. $\textbf{r}_e$ is the embedding of the relation $r_e$.

%\begin{align}
%    E_1(h,r,t)&=\Vert h+r-t \Vert \label{eq2}\\
%    E_2(p,r)&=R(p\vert h,t)(\prod_{\mu_i \in B(p)} \mu_i) \Vert C(p)-r \Vert \label{eq3}\\
%    E_3(r,r_e)&=\Vert r-r_e \Vert
%\label{eq4}
%\end{align}

\subsection{Objective Formalization}
\label{section3.4}

With the open world assumption \cite{Drumond:Tensor-Factorization}, we introduce the pairwise ranking loss function to formalize our optimization objective of RPJE for training, which is defined as
\small{
\begin{align}
    L= \sum_{(h,r,t)\in T} [&L_1(h,r,t)+\alpha_1 \sum_{p \in P(h,t)} L_2(p,r) \nonumber\\
      &+\alpha_2 \sum_{r_e \in D(r)} L_3(r,r_e)]
\label{eq5}
\end{align}}
\normalsize
In Eq.\ref{eq5}, $D(r)$ is defined as the set of relations all deduced from $r$ on the basis of Rules $R_1$, and $r_e$ is any relation in $D(r)$. $P(h,t)$ denotes all the paths linking entity pair $(h,t)$, and $p$ is one of the path in $P(h,t)$. $L_1(h,r,t)$, $L_2(p,r)$ and $L_3(r,r_e)$ are three margin-based loss functions considering the energy functions in Eqs.\ref{eq2},\ref{eq3},\ref{eq4} to measure the effectiveness of representation learning in regard to the direct triple $(h,r,t)$, the path pair $(p,r)$ as well as the relation pair $(r,r_e)$, respectively, which are defined as follows:
\begin{equation}
\resizebox{0.96\linewidth}{!}{$
    \displaystyle
    L_1(h,r,t)=\sum_{(h',r',t')\in T^-}max(0,\gamma_1+E_1(h,r,t)-E_1(h',r',t'))
    $}
\label{eq6}
\end{equation}

\begin{equation}
\resizebox{0.83\linewidth}{!}{$
    \displaystyle
    L_2(p,r)=\sum_{(r')\in T^-} max(0,\gamma_2+E_2(p,r)-E_2(p,r')
    $}
\label{eq7}
\end{equation}

\begin{equation}
\resizebox{0.9\linewidth}{!}{$
    \displaystyle
    L_3(r,r_e)=\sum_{(r')\in T^-} max(0,\gamma_3+\beta E_3(r,r_e)-E_3(r,r'))
    $}
\label{eq8}
\end{equation}
where the function $max(0,x)$ is defined to obtain the maximum value between 0 and $x$. $\gamma_1$, $\gamma_2$, $\gamma_3$ are three positive hyper-parameters denoting each margin of the loss functions in Eqs.\ref{eq6},\ref{eq7},\ref{eq8}, respectively. The weight of triples is fixed to 1, and $\alpha_1$, $\alpha_2$ are two hyper-parameters respectively weighting the influence of paths and relation pairs embedding constraint. $\beta$ denotes the confidence level of the rule in Rules $R_1$ associating $r$ and $r_e$. The confidence levels of all the rules are considered to be penalty coefficients in optimization. $T$ represents a set that contains all the positive triples observed in KG. Following the negative sampling method as in \cite{Bordes-Weston-Bengio:semantic-matching}, $T^-$ contains the negative triples reconstructed via randomly replacing the entities and relations in $T$ and removing the triples already exist in $T$.
\begin{align}
    T^-={(h',r,t)\cup(h,r',t)\cup(h,r,t')}
\end{align}

To solve the optimization, we utilize mini-batch stochastic gradient descent (SGD). And considering the training efficiency, the paths are limited no longer than 3 steps.

\section{Experiments}

\subsection{Experiment Settings}

\subsubsection{Datasets and Rules.}
We evaluate our model on four typical datasets: FB15K and FB15K-237 both extracted from the large-scale Freebase \cite{BGF:Freebase}, WN18 extracted from WordNet \cite{Miller:WordNet} and NELL-995 extracted from NELL \cite{Mitchell:nell}. Note that FB15K-237 contains no inverse relation and hence it is hard to learn embeddings by these mutually independent relations, so FB15K and FB15K-237 are always regarded as two distinguishing datasets. Statistics of datasets used are shown in Table \ref{table4}. We evaluate the performance of our approach and other baselines on KG completion task, which is specifically formulated as entity prediction and relation prediction. Specifically, entity prediction aims to complete a triple with one entity missing while relation prediction aims to predict a relation given head and tail entities.

\begin{table}\small
\centering
\begin{tabular}{cccccc}
\toprule
Dataset		& \#Rel		& \#Ent		& \#Train	& \#Valid	& \#Test\\
\midrule
FB15K		& 1,345		& 14,951	& 483,142	& 50,000	& 59,071\\
FB15K-237	& 237		& 14,541	& 272,115	& 17,535	& 20,466\\
WN18        & 18        & 40,943    & 141,442   & 5,000     & 5,000\\
NELL-995    & 200       & 75,492    & 123,370   & 15,000    & 15,838\\
\bottomrule
\end{tabular}
\caption{Statistics of datasets used in the experiments. Rel denotes relation and Ent denotes entity.}
\label{table4}
\end{table}

Our scheme is readily incorporable to any rule mining tool. And we choose AMIE+ \cite{Galarrage:AMIE} for its convenience and fast-speed to mine rich rules with an alternative confidence threshold on different databases. The confidence thresholds of rules are selected in the range of [0,1] with the step size 0.1 to search the best performance of rules on datasets. Table \ref{table5} lists the statistics of rules with various confidence thresholds in the range of $[0.5,0.9]$ mined from FB15K, FB15K-237, WN18 as well as NELL-995, which have been encoded for representation learning.

\newcommand{\tabincell}[2]{\begin{tabular}{@{}#1@{}}#2\end{tabular}}
\begin{table}\footnotesize
\centering
\begin{tabular}{ccccccc}
\toprule
\multirow{2}*{Datasets}			& \multirow{2}*{\tabincell{c}{Rule\\Types}}		& \multicolumn{5}{c}{Various Confidence Thresholds}\\
 &  & \#0.5		& \#0.6	& \#0.7		& \#0.8		& \#0.9\\
\midrule
%\multirow{2}*{FB15K}\\
\multirow{2}*{FB15K}		& $R_1$		& 1,157		& 975		& 899		& 767		& 586\\
			& $R_2$		& 643		& 632		& 586		& 535		& 229\\
\midrule
%\multirow{2}*{FB15K-237}\\
\multirow{2}*{FB15K-237}	& $R_1$		& 93		& 89		& 81		& 76		& 40\\
			& $R_2$		& 359		& 309		& 292		& 253		& 232\\
\midrule
%\multirow{2}*{WN18}\\
\multirow{2}*{WN18}	& $R_1$		& 17		& 17		& 17		& 17		& 16\\
			& $R_2$		& 89		& 80		& 77		& 24		& 0\\
\midrule
%\multirow{2}*{NELL-995}\\
\multirow{2}*{NELL-995}	    & $R_1$     & 132	    & 96		& 58		& 40		& 15\\
			                & $R_2$		& 326		& 266		& 201		& 161		& 105\\
\bottomrule
\end{tabular}
\caption{Statistics of the encoded rules in various confidence thresholds from the four datasets. Note that the rules in Rules $R_1$ extracted from WN18 have the nearly same amount for their confidence levels all exceed 0.8.}
\label{table5}
\end{table}

\subsubsection{Evaluation Protocols.}

Three principle assessment metrics are focused on: the mean rank of correct entities (MR), the mean reciprocal rank of correct entities (MRR) and the proportion of test triples for which correct entity is ranked in the top n predictions (Hits@n). And an evaluation result should achieve lower MR, higher MRR and Hits@10. Moreover, the “filtered” setting eliminates the reconstructed triples that could be observed in the KG, yet the “raw” setting does not. To achieve these metrics, We define the score function for calculating the scores for reconstructed triples as follows:
\begin{align}
    Q(h,&r,t) = \Vert \textbf{h}+\textbf{r}-\textbf{t} \Vert \nonumber\\
    		&+ \alpha_1 \sum_{p \in P(h,t)} R(p|h,t)(\prod_{\mu_i \in B(p)} \mu_i)\Vert C(p)-\textbf{r}\Vert
\label{eq10}
\end{align}
As shown in Eq.\ref{eq10}, the Rules $R_2$ should be utilized for composing paths in testing process. We rank the scores in descending order.

\subsubsection{Baselines for Comparison.}

To verify the performance of our approach, we select several involved state-of-the-art models to implement KG completion, including three types of baselines: (1) Embedding methods only considering triple facts: TransE \cite{Bordes:TransE}, TransH \cite{Wang:TransH}, TransR \cite{Lin-a:TransR}, STransE \cite{Nguyen:STransE}, TransG \cite{Xiao:TransG}, TEKE \cite{Wang:TextE}, R-GCN+ \cite{Michael:GNN}, KB-LRN \cite{Garciaduran:KBLRN}, ConvE \cite{Dettmers:CNN}. (2) Path-based models:  PTransE \cite{Lin-b:PTransE} and DPTransE \cite{DPTransE}. (3) Rule enhanced models: KALE \cite{Shu-Guo:Jointly} and RUGE \cite{Shu-Guo-2:RUGE}. We use the best results presented in their original papers and also implement PTransE and RUGE by their source codes.

\subsubsection{Experimental Settings.}

To guarantee fair comparison, we adopt the following evaluation settings in our work: (1) 100 mini-batches are created on datasets. (2) The entity and relation embeddings are initialized randomly and limited to unit vectors. (3) Following the same configurations as many prevailing baselines, the learning rate is chosen as 0.001, $\gamma_1$ and $\gamma_2$ are selected as $\gamma_1=\gamma_2=1$, the embedding dimension is set to 50 for WN18 and 100 for other three datasets considering only 18 relations exist in WN18, dissimilarity is selected as $L_1$ and training epochs is set to 500. In addition, we employ a grid search to select the other optimal hyper-parameters. We manually tune the margin $\gamma_3$ in $\{1, 1,5, 2, 2.5, 3\}$, and the weight coefficients $\alpha_1$, $\alpha_2$ both in $\{0,5, 1, 1.5, 2, 3, 5\}$. The best models are selected on validation sets. The resulting optimal of margin $\gamma_3$ and the weight coefficients $\alpha_1$, $\alpha_2$ are assigned to: $\gamma_3=1$, $\alpha_1=1$, $\alpha_2=3$.

\subsection{Influence of Confidence Levels and Path Steps}
\label{section4.5}

\begin{figure}
	\centering
	\includegraphics[width=1\columnwidth]{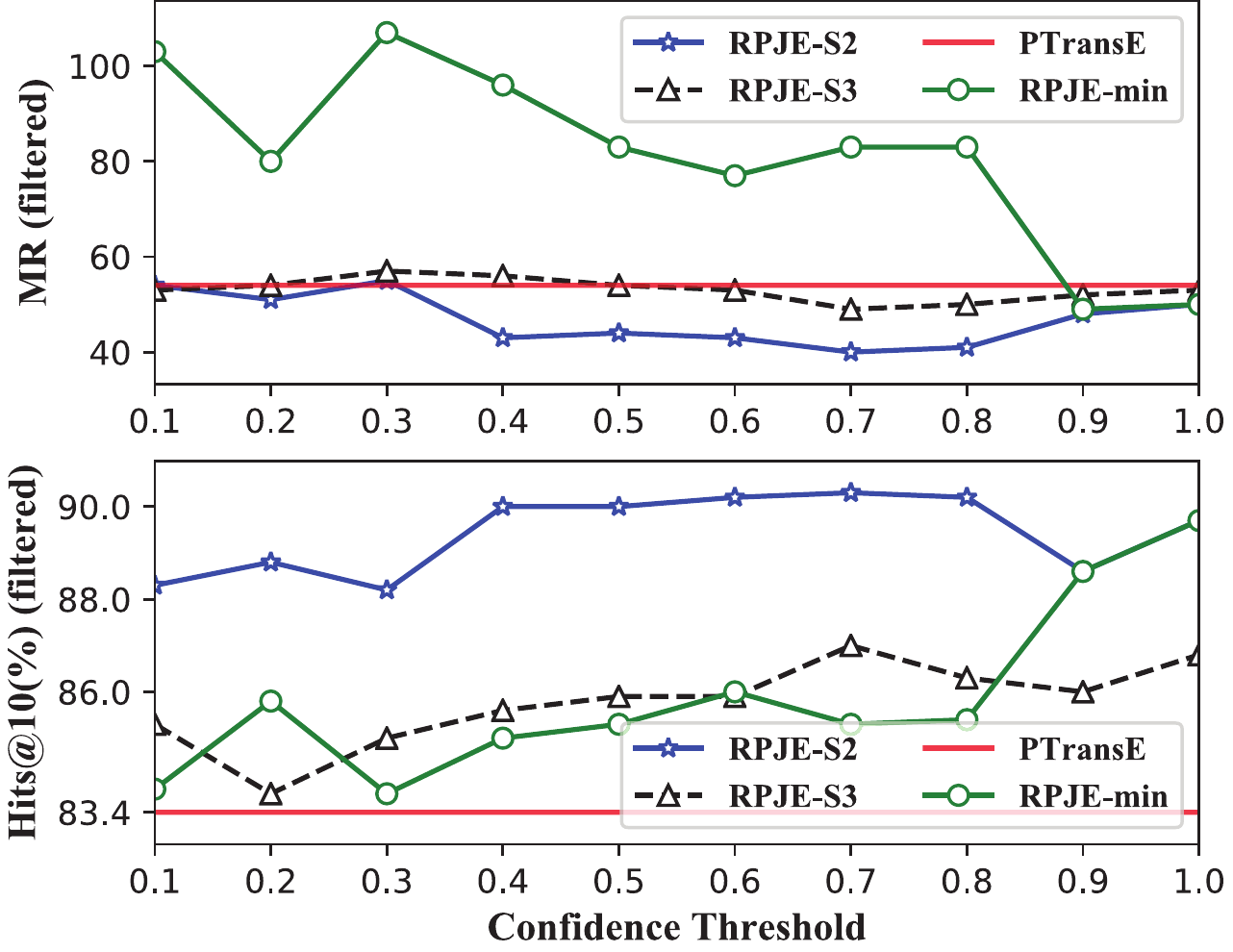}
	\caption{Performance comparison of various confidence thresholds and path steps limitation.}
	\label{figure3}
\end{figure}

In this subsection, we experimentally examine the impact of the two important parameters in our proposed scheme, namely the confidence levels of the rules and path steps. It indicates that our model is robust to the noisy rules for the rules with low confidence will be filtered out according to the appropriate confidence threshold. For instance, based on the confidence threshold 0.7, the rule $AthletePlaysInLeague(x,y) \Leftarrow AthleteledTeam(x,z) \wedge TeamPlaysInLeague(z,y)$ with confidence 0.96 will be used in path composition, but the rule $PersonHasJobPosition(x,y) \Leftarrow HasSibling(x,z) \wedge PersonHasJobPosition(z,y)$ with confidence 0.6 will be removed. It is also known that selecting a confidence threshold is a trade-off between higher confidence with more rules. We investigate the performance influence by varying the confidence thresholds in the range of [0.1, 1.0] with step 0.1. From Figure \ref{figure3}, we can observe that the confidence thresholds of 0.7 and 0.8 achieve the best tradeoffs. Furthermore, RPJE outperforms PTransE with the confidence threshold in a broad range of [0.4, 1.0] which illustrates the rules exploited in our model will be effective as long as the confidence threshold is selected in a moderate range. Particularly, RPJE-min obtains worse performance with lower confidence threshold due to more incorrect rules employed will cut down the accuracy of representation learning.

Additionally, we could compare the performance of different path steps limitation between steps 2 and 3. Figure \ref{figure3} illustrates the entity prediction results considering different confidence thresholds achieved by RPJE-S2 (2-step path), RPJE-S3 (3-step path), RPJE-min (RPJE ignoring the confidence of rules) and PTransE on FB15K. These results verify the confidence levels' contribution to representation learning. On account of same configurations, RPJE employing paths with maximum 2 steps consistently outperforms that with maximum 3 steps. The reason might be that longer paths may cause lower accuracy in paths composition, which will be studied in the future work. Therefore, we select the confidence threshold as 0.7 and the path steps as 2 for the best setting in the following results.

\begin{table}\small
\centering
\begin{tabular}{c|cc|c|cc}
\toprule
\multirow{2}*{Models} & \multicolumn{2}{|c|}{MR} & MRR & \multicolumn{2}{|c}{Hits@ 10(\%)}\\
	& raw	& filtered	& filtered	& raw	& filtered\\
\midrule
TransE	& 243		& 125	& 0.4		& 34.9		& 47.1\\
TransH	& 212		& 87	& -		& 45.7		& 64.4\\
TransR	& 198	& 77	& -			& 48.2		& 68.7\\
STransE	&219		&69		& 0.543		& 51.6		& 79.7\\
R-GCN+	& -			& -		& 0.696		& -			& 84.2\\
KB-LRN	&-			&\underline{44}		& \underline{0.794}		& -			& \underline{87.5}\\
ConvE	&-			& 51	& 0.657		& -			& 83.1\\
\midrule
PTransE		& 200	& 54	& 0.679		& 51.8	& 83.4\\
DPTransE	& \underline{191}	& 51	& -		& \textbf{58.1}	& 88.5\\
\midrule
KALE	& -			& -		& 0.523		& -			& 76.2\\
RUGE	& \underline{191}			& 71		& 0.768		& 54.3			& 86.5\\
\midrule
RPJE	& \textbf{186}	& \textbf{40}	& \textbf{0.816}	& \underline{55.0}	& \textbf{90.3}\\
\bottomrule
\end{tabular}
\caption{Entity prediction results on FB15K. The missing values indicate the scores not reported in the original work. The best score is in \textbf{bold}, and the second best in \underline{underline}.}
\label{table6}
\end{table}

\begin{table*}\small
\centering
\begin{tabular}{c|cccc|cccc}
\toprule
\multirow{2}*{Models}	& \multicolumn{4}{|c|}{Head Prediction (Hits@10)}	& \multicolumn{4}{|c}{Tail Prediction (Hits@10)}\\
	& 1-1	& 1-N	& N-1	& N-N	& 1-1	& 1-N	& N-1	& N-N\\
\midrule
TransE	& 35.6	& 62.6	& 17.2	& 37.5	& 34.9	& 14.6	& 68.3	& 41.3\\
TransH	& 66.8	& 87.6	& 28.7	& 64.5	& 65.5	& 39.8	& 83.3	& 67.2\\
TransR	& 78.8	& 89.2	& 34.1	& 69.2	& 79.2	& 37.4	& 90.4	& 72.1\\
STransE	& 82.8	& 94.2	& 50.4	& 80.1	& 82.4	& 56.9	& 93.4	& 83.1\\
TEKE	& 78.8	& 89.3	& 54.0	& 81.7	& 79.2	& 59.2	& 90.4	& 83.5\\
TransG	& \underline{93.0}	& \underline{96.0}	& \underline{62.5}	& \underline{86.8}	& 92.8	& 68.1	& \underline{94.5}	& \underline{88.8}\\
PTransE	& 91.0	& 92.8	& 60.9	& 83.8	& 91.2	& \underline{74.0}	& 88.9	& 86.4\\
DPTransE & 92.5	& 95.0	& 58.0	& 86.6	& \underline{93.5}	& 71.1	& 93.9	& 88.2\\
\midrule
RPJE	& \textbf{94.2}	& \textbf{96.5}	& \textbf{70.4}	& \textbf{91.6}	& \textbf{94.1}	& \textbf{83.9}	& \textbf{95.3}	& \textbf{93.3}\\
\bottomrule
\end{tabular}
\caption{Entity prediction results on FB15K by mapping properties of relations (\%) with filtered setting.}
\label{table7}
\end{table*}

\begin{table}\small
\centering
\begin{tabular}{c|cc|cc}
\toprule
\multirow{2}*{Models} & \multicolumn{2}{|c|}{MR}	& \multicolumn{2}{|c}{Hits@1(\%)}\\
		& raw		& filtered		& raw		& filtered\\
\midrule
TransE	& 2.79		& 2.43			& 68.3		& 87.2\\
PTransE	& 1.81	& 1.35			& 69.5		& 93.6\\
RUGE	& 2.47	& 2.22			& 68.8		& 87.0\\
\midrule
RPJE	& \textbf{1.68}		& \textbf{1.24}			& \textbf{70.1}		& \textbf{95.3}\\
\bottomrule
\end{tabular}
\caption{Relation prediction results on FB15K. We use Hits@1 for better comparison because Hits@10 of all the models exceed 95\%.}
\label{table8}
\end{table}

\subsection{Evaluation Results on FB15K and FB15K-237}

In this section, we first evaluate entity prediction and relation prediction of the proposed RPJE with a variety of baselines on FB15K. From Table \ref{table6}, it can be observed that: (1) Our approach RPJE achieves superiority compared with other baselines, and most of the improvements are statistically significant. This demonstrates that RPJE learns more reasonable embeddings for KGs via using logic rules in conjunction with paths. (2) In particular, RPJE outperforms PTransE on each metric, which indicates the superiority of introducing logic rules for providing higher accuracy in paths composition and learning better path embeddings. (3) Compared to the rule-based baselines KALE and RUGE, RPJE obtains the improvements of 56.0\%/6.3\% on MRR and 18.5\%/4.4\% on Hits@10 (filtered), which demonstrates the effectiveness of explicitly employing rules for preserving more semantic information and further integrating paths.

Table \ref{table7} shows the evaluation results of predicting entities by various types of relations. We can observe that: 1) RPJE outperforms all baselines significantly and consistently in regard to all the relation categories. Compared to the best performing baseline TransG, RPJE achieves an average improvement of 4.2\% in head entities prediction and 6.5\% in tail entities prediction. 2) More interestingly, on the two toughest tasks of predicting head entities of N-1 relation and predicting tail entities of 1-N relation, our approach achieves the best performance improvements approximately 12.6\% and 13.4\% compared to the best baselines, respectively.

The results of relation prediction are shown in Table \ref{table8}. Three typical models representing three types of baselines are implemented. The results illustrate that RPJE outperforms baselines in all metrics. It verifies that paths could provide extra relationships for entity pairs and rules can further create more semantic association for relations to improve relation embeddings and benefit for relation prediction.

Furthermore, we implement the experiments on dataset FB15K-237. Since FB15K-237 is constructed up to date, only a minority of existing works have implemented their experiments and show evaluation results on this dataset, which can be selected as baselines. As shown in Table \ref{table9}, RPJE obtains the best performance with approximately 29.5\% improvement compared to PTransE on MRR and 26.8\% improvement compared to KB-LRN on Hits@10. Although no inverse relation could be observed in FB15K-237, we could employ Horn rules to provide significant supplements for building semantic associations of relations.

\begin{table}\small
\centering
\begin{tabular}{c|c|c|c}
\toprule
Models  & MR	& MRR	& Hits@10(\%)\\
\midrule
TransE	& 347		& 0.294			& 46.4\\
R-GCN+	& -			& 0.249			& 41.7\\
KB-LRN	& \underline{209}		& 0.309			& 49.3\\
ConvE	& 246		& 0.316			& 41.7\\
\midrule
PTransE	& 302	& \underline{0.363}			& \underline{52.6}\\
RUGE	& 488	& 0.164			& 34.9\\
\midrule
RPJE	& \textbf{207}		& \textbf{0.470}			& \textbf{62.5}\\
\bottomrule
\end{tabular}
\caption{Entity prediction results on FB15K-237.}
\label{table9}
\end{table}

\begin{table}\small
\centering
\begin{tabular}{c|cc|cc}
\toprule
\multirow{2}*{Models} & \multicolumn{2}{|c|}{WN18}	& \multicolumn{2}{c}{NELL-995}\\
		        	& MRR		& Hits@10(\%)		& MRR		& Hits@10(\%)\\
\midrule
TransE              & 0.495		& 93.4              & 0.219     & 35.2\\
PTransE             & 0.890		& 94.5              & 0.304     & 43.7\\
RUGE             & 0.943		& 94.4              & 0.318     & 43.3\\
\midrule
RPJE	            & \textbf{0.946}	& \textbf{95.1}     & \textbf{0.361}  & \textbf{50.1}\\
\bottomrule
\end{tabular}
\caption{Entity prediction results on WN18 and NELL-995.}
\label{table10}
\end{table}

\subsection{Evaluation Results on WN18 and NELL-995}

We also test the models on datasets WN18 and NELL-995. Three types of typical models are selected as baselines. Few rules can be mined from WN18 due to extremely limited amount of relations. And very parse paths can be extracted from NELL-995 because entities are far more than relations on this dataset. Even so, our model RPJE achieves consistent and significant improvements over other baselines as shown in Table \ref{table10}. It illustrates the superiority of our approach for representation learning on various large scale KGs. On the other hand, considering the performance gains on FB15K are more than that on WN18, which is because more rules provide more semantic information to RPJE to use. As can be expected, our model RPJE will obtain better performance on datasets which implies more rules and paths.

\subsection{Ablation Study}
To verify the effectiveness of different components of RPJE, we implement the ablation study of entity prediction on FB15K by removing the paths as well as the length-2 rules at one time (-PaRu2), and the rules of length 1 (-Ru1) from our integrated model, respectively. More specifically, -PaRu2 means removing $E_2$/$L_2$ in Eq. \ref{eq3}/\ref{eq7} and -Ru1 means removing $E_3$/$L_3$ in Eq. \ref{eq4}/\ref{eq8}. As shown in Table \ref{table11}, we can conclude removing each component will lead to performance degradation especially when removing the paths and the rules of length 2 changes the results significantly.

\begin{table}\small
\centering
{
\begin{tabular}{c|cc|c|cc}
\toprule
\multirow{2}*{Models} & \multicolumn{2}{|c|}{MR} & MRR & \multicolumn{2}{|c}{Hits@ 10(\%)}\\
	& raw	& filtered	& filtered	& raw	& filtered\\
\midrule
RPJE	            & \textbf{186}      & \textbf{40}     & \textbf{0.816}    & \textbf{55.0}    & \textbf{90.3}\\
-PaRu2              & 205               & 63		      & 0.453             & 49.3             & 72.1\\
-Ru1                & 193               & 47		      & 0.812             & 54.5             & 90.0\\
\bottomrule
\end{tabular}}
\caption{Ablation study by removing paths and length-2 rules as well as length-1 rules.}
\label{table11}
\end{table}

\subsection{Case Study}

\begin{figure}
	\centering
	\includegraphics[width=1\columnwidth]{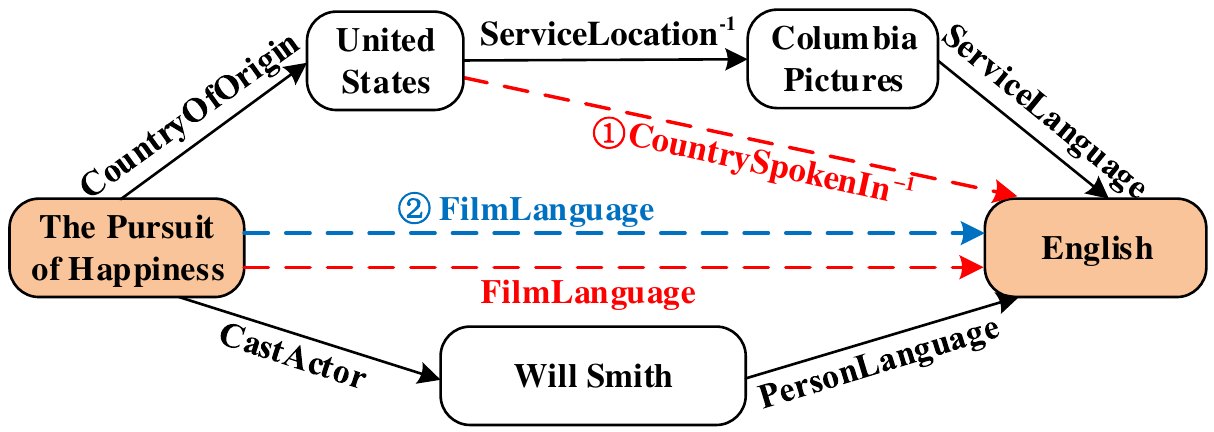}
	\caption{An example of the explainable relation prediction.}
	\label{figure4}
\end{figure}

As shown in Figure \ref{figure4}, considering a relation prediction task with the given head entity \small$The Pursuit of Happiness$ \normalsize and tail entity \small$English$\normalsize, the result \small$FilmLanguage$ \normalsize is obtained by our model RPJE. Particularly, this result can be explained by RPJE with paths and rules: for the 2-steps path, the rule
\begin{equation}\nonumber
  \resizebox{1\columnwidth}{!}{$
  filmlanguage(x,y)\Leftarrow castactor(x,z)\wedge personlanguage(z,y)
  $}
\end{equation}
with the confidence 0.81 is activated to compose the path into the prediction result \small$FilmLanguage$ \normalsize while providing the confidence level 0.81 of this result. For the path of 3 steps, the intermediate composition result \small$CountrySpokenIn^{-1}$ \normalsize is obtained by calling Rules $R_2$ and further employed to achieve the relation prediction result $FilmLanguage$ via embeddings on the reconstructed path containing the relation \small$CountrySpokenIn^{-1}$\normalsize.

%s shown in Figure \ref{figure4}, the relation prediction result $FilmLanguage$ is obtained to link the entity pair $The Pursuit of Happiness$ and $English$ by our model RPJE. What's more, we can show the explanation for this result. For the path of 3 steps, we first infer the relation $CountrySpokenIn^{-1}$ by Rules $R_2$ and further achieve the relation prediction result of the whole path via embeddings. And for the path of 2 steps, the relation prediction result could directly obtained by Rules $R_2$. Obtaining the same inferred relation $FilmLanguage$ based on two paths could demonstrate more confidence of the results.

\section{Conclusion and Future Work}

In this paper, we proposed a novel model RPJE to learn KG embeddings by integrating triple facts, Horn rules and paths in a unified framework to enhance the accuracy and the explainability of representation learning. The Experimental results on KG completion verified the rules with confidence levels are significant in improving the accuracy of composing paths and enhancing the association between relations.

For future work, we will investigate some other potential composition operations such as Long Short Term Memory networks (LSTM) with attention mechanism which may benefit for long paths. And we will explore to push the embedding information back from RPJE to rule learning with a well-designed closed-loop system.

\section{Acknowledgements}

This work was partially supported by the National Natural Science Foundation of China (No. 61772054), the NSFC Key Project (No. 61632001), the Natural Science Foundation of China (No. 61572493) and the Fundamental Research Funds for the Central Universities.

\small
\bibliography{aaai-2020}
\bibliographystyle{aaai}
\end{document}